\documentclass[sigconf]{acmart}

\usepackage{booktabs} 

\usepackage{graphicx}
\usepackage{subfigure}
\usepackage{multirow}

\usepackage{algorithm}
\usepackage{algpseudocode}
\usepackage{amsmath}
\usepackage{graphics}
\usepackage{epsfig}
\usepackage{enumitem}

\acmArticle{4}
\acmPrice{15.00}

\begin{document}

\title{Image Matters: Visually modeling user behaviors using Advanced Model Server}

\author{
Tiezheng Ge, Liqin Zhao, Guorui Zhou, Keyu Chen, Shuying Liu \\
Huimin Yi, Zelin Hu, Bochao Liu, Peng Sun, Haoyu Liu, Pengtao Yi, Sui Huang \\  
Zhiqiang Zhang, Xiaoqiang Zhu, Yu Zhang, Kun Gai \\
 Alibaba Inc. \\
 \{tiezheng.gtz, zhang.zhiqiang, jingshi.gk\}@alibaba-inc.com}

\renewcommand{\shortauthors}{T. Ge et al.}

\begin{abstract}

In Taobao, the largest e-commerce platform in China, billions of items are provided and typically displayed with their images.
For better user experience and business effectiveness, Click Through Rate (CTR) prediction in online advertising system exploits abundant user historical behaviors to identify whether a user is interested in a candidate ad. 
Enhancing behavior representations with user behavior images will help understand user's visual preference and improve the accuracy of CTR prediction greatly.
So we propose to model user preference jointly with user behavior ID features and behavior images. 
However, training with user behavior images brings tens to hundreds of images in one sample, giving rise to a great challenge in both communication and computation. 
To handle these challenges, we propose a novel and efficient distributed machine learning paradigm called Advanced Model Server (AMS).
With the well-known Parameter Server (PS) framework, each server node handles a separate part of parameters and updates them independently.
AMS goes beyond this and is designed to be capable of learning a unified image descriptor model shared by all server nodes which embeds large images into low dimensional high level features before transmitting images to worker nodes.
AMS thus dramatically reduces the communication load and enables the arduous joint training process. 
Based on AMS, the methods of effectively combining the images and ID features are carefully studied, and then we propose a Deep Image CTR Model. Our approach is shown to achieve significant improvements in both online and offline evaluations, and has been deployed in Taobao display advertising system serving the main traffic.

\end{abstract}


%
%

\fancyhead{}
\settopmatter{printacmref=false}
\renewcommand\footnotetextcopyrightpermission[1]{}
\pagestyle{plain}
\copyrightyear{2018} 
\acmYear{2018} 
\setcopyright{acmcopyright}
\acmConference{KDD'18}{}{August 19--23, 2018, London, United Kingdom.}




\maketitle

\vspace{-0.2cm}
\section{Introduction}

 



\begin{figure}
\subfigure[]{
\label{display_advertising} 
\includegraphics[width=0.245\linewidth]{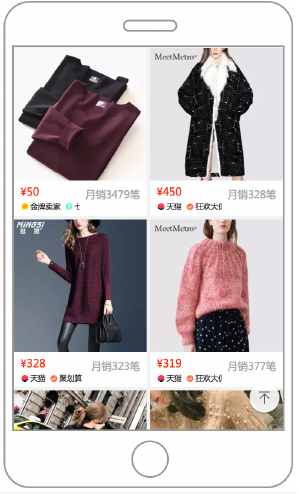}}
\subfigure[]{
\label{advertising_pipeline} 
\includegraphics[width=0.715\linewidth]{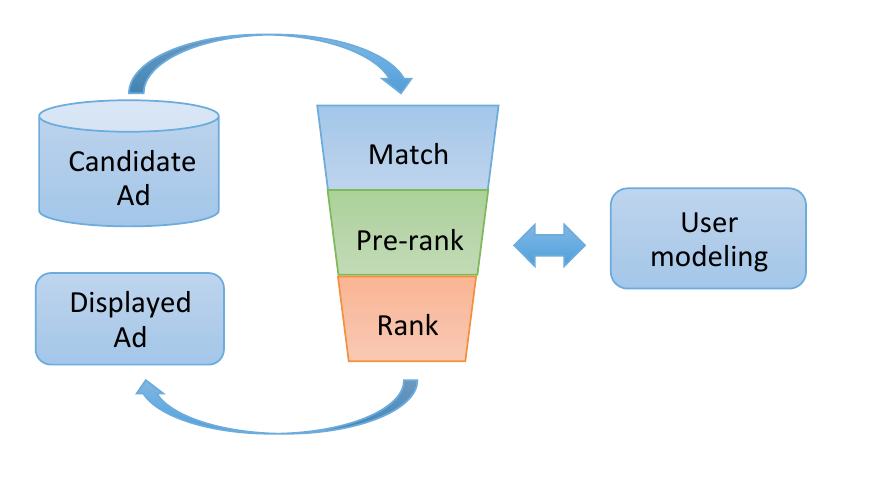}}
\vspace{-0.4cm}
\caption{(a) Typical display ad on App of Taobao. (b) Display advertising pipeline}
\vspace{-0.3cm}
\label{fig:gauc} 
\end{figure}

Taobao is the largest e-commerce platform in China, serving hundreds of millions of users with billions of items through both mobile app and PC website.
Users come to Taobao to browse these items through the search or personalized recommendation.
Each item is usually displayed by an item image along with some describing texts.
When interested in an item, users click that image to see the details.
Fig~\ref{display_advertising} shows an example of recommended items in Taobao mobile app.

Taobao also established one of the world's leading display advertising systems, helping millions of advertisers connect to customers.
By identifying user interests, display ads are presented in various spots like Guess What You Like and efficiently deliver marketing messages to the right customers.
Cost-per-click (CPC) pricing method is adopted and sufficiently effective~\cite{Zhu2017Optimized}.
In CPC mode, the ad publishers rank the candidate ads in order of effective cost per mille (eCPM), which can be estimated by the product of the bid price and the estimated click through rate (CTR).
Such strategy makes CTR prediction the core task in the advertising system.





CTR prediction scores a user's preference to an item, and largely relies on understanding user interests from historical behaviors.
Users browse and click items billions of times in Taobao every day, and these visits bring a huge amount of log data which weakly reflect user interests.
Traditional researches on CTR prediction focus on carefully designed feedback feature~\cite{yahoo1,yahoo2} and shallow models, \emph{e.g.}, Logistic Regression~\cite{richardson2007predicting}.
In recent years, the deep learning based CTR prediction system emerged overwhelmingly~\cite{zhang2017deep}.
These methods mainly involve the sparse ID features, \emph{e.g.}, ad ID, user interacted item ID, \emph{etc}.
However, when an ID occurs less frequently in the data, its parameter may not be well trained.
Images can provide intrinsic visual descriptions, and thus bring better generalization for the model.
Considering that item images are what users directly interact with, these images can provide more visual information about user interests.
We propose to naturally describe each behavior by such images, and jointly model them with ID features in CTR prediction.


Training CTR models with image data requires huge computation and storage consumption.
There are pioneering works~\cite{chen2016deep,mo2015image} dedicating to represent ad with image features in CTR prediction.
These studies did not explore user behavior images.
Modeling user behavior images can help understand user visual preference and improve the accuracy of CTR prediction.
Moreover, combining both user visual preference and ad visual information could further benefit CTR prediction.
However, modeling user preference with interacted images is more challenging.
Because the number of one typical user's behaviors ranges from tens to hundreds, it will bring the same number of times the consumption than that when only modeling ad images, and is a non-trivial problem. A well designed efficient training system is a must to handle this large scale problem for real production.

We propose Advanced Model Server (AMS) framework, which goes beyond the well-known Parameter Server (PS)~\cite{smola2010architecture, li2014scaling} to handle this large scale training problem.
The key motivation is to learn a unified high level image descriptor among all server nodes, which embeds large images into low dimensional features before transmitting raw images to workers. In this way, not only can the communication load be largely reduced, but also the repetitive computation of descriptor can be aggregated in servers. With traditional PS framework, however, each of the server nodes handles a separate part of parameters and updates them independently.
Thus PS lacks the ability to learn the unified image descriptor which is shared among all server nodes.
In AMS, servers are designed to be capable of forwarding and globally updating a shared learnable sub-model with images distributed on each server node.

With this design, the whole CTR model is divided into the worker model, which assembles all features to predict CTR on workers, and the server model, which is the high level image descriptor learned on servers.
Then, raw image feature data are distributed on server side as globally shared features without repetition, largely reducing storage usage than the in-sample storage, in our application, by about 31 times.
And only low dimensional high-level semantic representations of images output by the server model part rather than raw images need to be transmitted, dramatically reducing communication load, in our application, by about 32 times.
Moreover, gradients are back propagated completely from worker model to server model, which guarantees an end-to-end training from raw image features to the final CTR score.

Based on AMS, we successfully build a highly efficient training system and deploy a light weight online service, which manages to tackle the heavy load of storage, computation and communication brought by the image features. Specifically, our training process with billions of samples finishes in 18 hours, enabling the daily update of online model, a required character for industrial production.

Benefitting from the carefully optimized infrastructure, we propose a unified network architecture, named Deep Image CTR Model (DICM) that effectively models users with behavior images.
DICM achieves image aware user modeling through a selected attentive pooling scheme, which employs both images and ID features in generating attention weights.
DICM also utilizes the visual connections between user preferences and ads, improving the performance remarkably.



To summarize, our contributions are three folds:

First, we propose the novel AMS framework.
It goes beyond the well-known parameter distributing style with a sub-model distributing style, and facilitates the jointly learning of the whole model in a distributed manner.
This is an important step towards enabling deep learning model to exploit large-scale and structured data with affordable computation and storage resources.

Second, we propose the DICM. 
It not only models ad with its image, but also exploits user's massive behavior images to better model user preference, which is much more challenging than that only uses ad images. 
We show that either ad images or user behavior images can benefit CTR prediction, and their combination will bring further significant improvement.

Moreover, we validate the efficacy and efficiency of our approach with extensive offline and online experiments. 
It has been now deployed in Taobao's display advertising system, serving the main traffic for half a billion users and millions of advertisers.




\section{Related work}


Early CTR prediction focuses on carefully designed low-dimensional statistical features, normally defined by votes of users' clicks, \emph{etc.}~\cite{yahoo1,yahoo2}.
LS-PLM~\cite{gai2017learning}, FTRL~\cite{FTRL} and FM~\cite{rendle2010factorization} are classical explorations on shallow models.
Recently, along with the number of samples and the dimension of features become larger and larger, CTR models evolve from shallow to deep.
Especially, inspired by natural language process field, the embedding technique which learns distributed representations is used to handle large scale sparse data.
NCF~\cite{He2017neural} and Wide\&Deep~\cite{WideandDeep} exploit MLP network to greatly enhance model capacities.
DeepFM~\cite{guo2017deepfm} further models feature interactions by updating the wide part with factorization machine in Wide\&Deep.
The latest work DIN~\cite{zhou2017deep} proposes to apply attentive mechanism to adaptively model user behaviors according to a given item.
These work have advanced the employment of sparse features.
However, IDs only tell objects are different, and reveal little semantic information.
Especially when an ID occurs in low frequency in training data, its parameters will not be well trained.
And unseen ID during training will not take effect in prediction.
Images with visual semantic information would bring better generalization ability of models.
Further, unseen images in training data can still help CTR prediction with a well trained image model.



Image representation task makes a significant improvement in recent years.
High level semantic features learnt by deep models~\cite{alexnet,Simonyan2014Very,szegedy2015going,he2016deep} have been proved to be effective on a large range of tasks.
Some previous works try to introduce image information in CTR model to describe ads.
Cheng et al.~\cite{cheng2012multimedia} and Mo et al.~\cite{mo2015image} address the cold start problem by modeling ad images with either manually designed feature or pre-trained CNN models.
Lynch et al. ~\cite{lynch2016images} introduce the visual information of items to overcome the misunderstanding of text-only representation in Esty's search engine. Chen et al.~\cite{chen2016deep} propose to train the CNN in an end-to-end manner.
All these works focus on representing ads with images, which differs from our motivation.
Images of ads describe visual features of ads, and user behavior images will reveal visual preferences of users.
Combining them together and bridging these visual information would result in better performance than either of them alone.
In this paper, we propose to enhance the user representation with images and design a novel and efficient distributed machine learning paradigm to tackle the challenges brought by it.

\begin{figure*}
  \centering
  \includegraphics[width=0.95\linewidth]{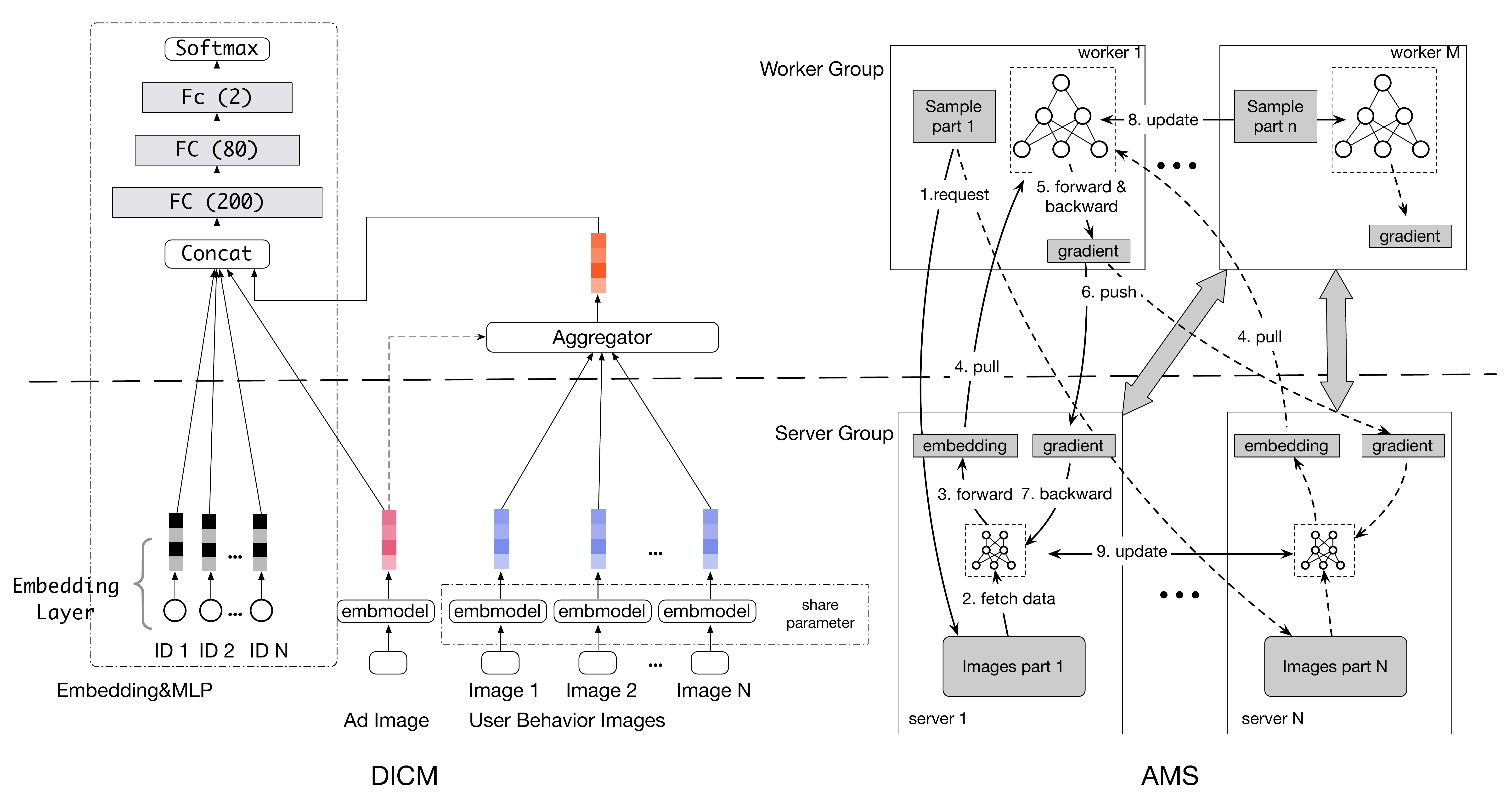}\\
  \vspace{-0.3cm}
  \caption{DICM network architecture implemented by Advance Model Server}\label{Networks}
  \vspace{-0.1cm}
\end{figure*}

\section{Deep Image CTR Model}

\subsection{Display advertising system} \label{ch_display_adv_sys}

Taobao's display advertising system responds to billions of page view (PV) requests everyday.
For each request, the most suitable ad is shown to a specific user in a specific scenario (viewing time, advertising position, \emph{etc.}).
The advertising system chooses the one ranked highest under eCPM mechanism from tens of millions of ads within only tens of milliseconds.

The online system completes this task in a funnel-like way and roughly consists of three sequential modules, as shown in Fig.~\ref{advertising_pipeline}.
The \emph{matching} module retrieves roughly around 4k ads from all candidates by current user's preferences inferred from its behaviors.
The successive \emph{Pre-rank} module further shrinks the number of candidates to about 400 with a light-weight CTR model.
Finally, the \emph{Rank} module predicts CTR of ads accurately with sophisticated models and ranks them by eCPM to give the best choice.
All of these modules depend on the appropriate understanding of user interests to give personalized recommendations.
Fig.~\ref{display_advertising} shows a typical advertising result in Taobao mobile app.




In this paper, we focus on better user modeling with user behavior images in CTR prediction. In the following sections, we carefully describe the challenges and solutions by  taking \emph{Rank} as an example. We also apply it to \emph{Pre-rank}. Both results for \emph{Rank} and \emph{Pre-rank} are shown later.
Our approach can also be used with tree based deep model~\cite{zhu2018learning} in \emph{Matching}, and we leave this as future work.

\subsection{Problem Formulation}\label{ch_problem}

Being fed features about user, ad and scenario \emph{etc.}, CTR model outputs a probability that the ad will be clicked by the user in that scenario.
Following previous works~\cite{covington2016deep,chen2016deep,mo2015image,zhou2017deep}, it is considered as a binary classification problem, whose label is the weak feedback---being clicked or not.
And the cross-entropy error~\cite{de2005tutorial} is used as the objective function during training.



Different from classifying images which are well represented by pixels, CTR prediction problem needs carefully feature design with respect to the specific application.
The common practice is to describe various aspects of user, item and scenario in each sample with underlying IDs, constituting many sparse feature fields.
User historical behavior field consists of the ids of items the user previously clicked and is the most important one describing users.
Such method leads to large scale but extremely sparse data.



Embedding with MLP fashion networks~\cite{He2017neural, WideandDeep, guo2017deepfm} are now widely used to fit such large sparse inputs.
In Taobao's advertising system, highly optimized CTR models following this pattern are deployed.
Embedding\&MLP part of Fig.~\ref{Networks} shows a simplified version of the production model for clarity.
Recently DIN~\cite{zhou2017deep} is introduced in production to better model sparse behavior features.
Working with these sophisticated models, in the following sections we show that modeling user behavior with images can still bring significant improvements.



\vspace{-0.03cm}
\subsection{Modeling with image}



We extend Embedding\&MLP model with visual information, especially enhance the user behavior representations with images.
We refer to this structure as Deep Image CTR Model (DICM) and refer to Embedding\&MLP as the basic net.
As illustrated in Fig.~\ref{Networks}, users' behavior images and ad images are incorporated as two special feature fields.
These images are first fed through a trainable sub-model to get high level representations with low dimensionality.
Similar to embedding, the sub-model is also a kind of embedding operation which embeds image to vector.
So we call it embedding model.
The embedding model can be regarded as a generalizable extension of traditional key-value embedding operation, for it can embed new images unseen during training.
Since user behaviors are of variable quantity, multiple embedded images need to be aggregated together into a fixed-length user representation,
and then be fed into MLP.

It is worth noting that image embedding in this model is actually independent, \emph{i.e.}, it does not depend on other features.
Thus the embedding model can be forward/backward separately.
This observation prompts us to design the Advanced Model Server.
Moreover, more embedding models for various types of data, \emph{e.g.} text, videos, can be devised using AMS.

\section{Advanced Model Server}

The main challenge of training is the huge quantity of images involved in user behavior.
The image not only is large-size datasource itself, but also needs complex computation in extracting semantic information.
For CTR prediction, each sample contains a user description with its massive historical behaviors.
Therefore, the training system inevitably faces the heavy load of storage, computation and communication.
For example, a typical user would have over 200 behaviors during our statistic period, which means a single training sample will involve over 200 images, hundreds of times more than that merely employs ad image.
Moreover the training system needs to handle billions of training samples and finish model update daily, which is required for online production.


Advanced Model Server (AMS) provides an efficient distributed training paradigm by introducing a shared and learnable sub-model on servers which significantly reduces the size of features to be transmitted.
AMS goes beyond the classical Parameter Server ~\cite{smola2010architecture, li2014scaling} in the sense that AMS is of capable of forwarding and updating this sub-model shared by all server nodes.




\begin{algorithm}[htb] 
\caption{Advanced Model Server} 
\label{alg:AMS} 
\begin{algorithmic}[1] 

\renewcommand{\algorithmicensure}{\textbf{Task Scheduler:}} 
\Ensure 
\State Initialize worker models $\mathcal{W}$ and embedding models $\mathcal{E}$
\For{Train with mini-batch t = 0,...,T}

\State do WORKERITERATION(t) on all workers.
\EndFor

\renewcommand{\algorithmicensure}{\textbf{Worker: r = 1,...,M}} 
\Ensure 
\Function{WORKERITERATON}{t}

\State load training data of mini-batch t: $X_r^t, Y_r^t$



\State request SERVEREMBED with image indices in $X_r^t$

\State pull all embeddings $e_r^t$ from servers

\State forward and backward $\mathcal{W}$ with $e_r^t$

gradient \emph{w.r.t} worker param $\delta_{\mathcal{W}_r}^t$ = $\nabla_w \ell(X_r^t, Y_r^t, e_r^t)$

gradient \emph{w.r.t} embeddings $\delta_{e_r}^t$ = $\nabla_e \ell(X_r^t, Y_r^t, e_r^t)$

\State push $\delta_{e_r}^t$ to servers' SERVERUPDATE


\State synchronize $\delta_{\mathcal{W}_r}^t$ with all workers and update $\mathcal{W}$


\EndFunction

\renewcommand{\algorithmicensure}{\textbf{Server: s = 1,...,N}} 
\Ensure 

\Function{SERVEREMBED}{t}

\State get image data $I$ from local

\State compute embeddings $e$ = $\mathcal{E}(I)$



\EndFunction

\Function{SERVERUPDATE}{t}

\State compute gradients \emph{w.r.t} embedding model at server \textbf{s}

$\delta_{\mathcal{E}_s}^t$ = $\nabla \mathcal{E}(I) \cdot \delta_{e}^t$

\State synchronize $\delta_{\mathcal{E}_s}^t$ with all servers and update $\mathcal{E}$



\EndFunction
\end{algorithmic} 
\end{algorithm}

\subsection{From Parameter Server to AMS}

Parameter Server (PS) is a widely adopted distributed architecture for large scale parametric machine learning problem. It consists of two node groups: the \emph{worker group} and the \emph{server group}. The \emph{worker group} contains a set of workers which do training on its assigned part of training samples. Meanwhile, the \emph{server group} serves as a distributed database which stores parameters of the models and can be accessed through key-value structure. 
In this way, PS store and maintain a large parameter set in server without redundancy.

The Embedding\&MLP model can be efficiently implemented with the PS-like architecture on GPU cluster. The parameters of embedding layer are placed in \emph{server group} since their size far exceeds the memory capacity of each worker, and can be accessed (forward) and updated (backward) through key-value structure.

However, when image features, especially the tremendous user behavior related images, are employed, to complete the training procedure is not trivial.


Actually if images are stored in \emph{worker group} along with training samples, then image features will greatly increase the training data size (in our scenario, from 134M Bytes to 5.1G per mini-batch, about 40 times larger), which makes it unaffordable for IO or storage.

And since the image feature is high-dimensional (typically 4096-D in our experiments) and far beyond that of ID features (typically 12-D), if images are stored in \emph{server group} and accessed by workers during training, then it will bring heavy communication pressure.

To handle this issue, the key motivation of AMS is that by learning a unified embedding model in servers, the high dimensional raw image features can be embedded into low dimensions before being transmitted to workers.

\begin{figure*}[htp]
  \centering
  \includegraphics[width=0.95\linewidth]{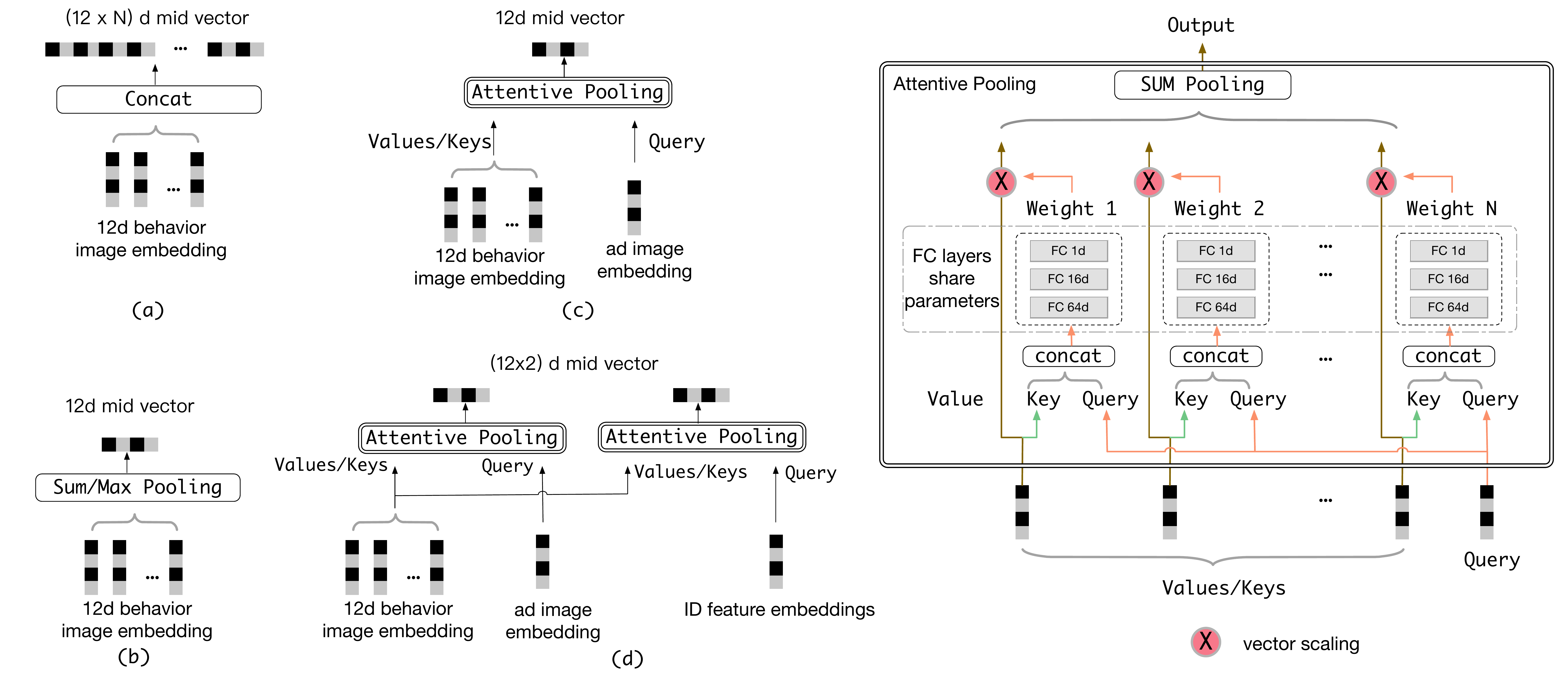}
  \vspace{-0.3cm}
  \caption{Aggregator architectures. (a) Concatenate (b) Sum/Max Pooling (c) Attentive Pooling (d) MultiQuery-AttentivePooling}\label{Aggregator}
\end{figure*}

\subsection{AMS architecture}

In this section, we detail the architecture of AMS, as illustrated in Fig.~\ref{Networks} and Algorithm~\ref{alg:AMS}.
AMS includes two parts of nodes, i.e. servers and workers.
Training samples are distributed among all workers and 
only image indices are contained in the samples for image features.
Image data are evenly distributed among all servers without duplicates in key-value format, in which the key is an image index and the value is its image data.


In each iteration, each worker node first reads a mini-batch of samples separately and then requests the image data with image indices occurred in the samples from server nodes. 
Note that image data are evenly distributed on servers, requests from different workers for the same image will be gathered by the same server and processed only once.
When receiving the request, the server first fetches the image data from local memory and then feeds it through the embedding model $\mathcal{E}$ to get an embedded vector $e$. 
Then each worker pulls back embedded vector $e$ instead of raw image data from servers to complete worker model computation and obtain gradients $\delta_{\mathcal{W}_r}$ and $\delta_{e_r}$ for worker model and embeddings $e$ respectively.
Once again, $\delta_{e_r}$ for the same image are collected by the same server and they are then back propagated through the embedding model to compute the gradient $\delta_{\mathcal{E}_s}$ \emph{w.r.t} embedding model at server \textbf{s}.
Finally workers and servers synchronize and accumulate their model gradients $\delta_{\mathcal{W}_r}$ and $\delta_{\mathcal{E}_s}$ and finish the model update.

Our AMS should be distinguished from PS. By the design of PS, parameters are distributed stored in \emph{server group}, and servers are mutually independent without communication. While in AMS, image data are distributed stored and parameters of embedding model are globally shared. In other words, the parametric model in servers is essentially unified and is distributed trained by all servers. The gradient synchronization among servers in each training iteration is critical in AMS. Note that although PS supports user defined functions in servers~\cite{li2014scaling}, these functions are fixed and are not trainable.


AMS brings several benefits.
First, the storage of images is significantly reduced by only storing once in servers.
Further the communication is reduced for the embedding vectors are much smaller than original data (typically from 4096-D to 12-D, over 340 times compression ratio).
Another benefit is that the computation of a certain image occurred several times within one training iteration can be naturally merged by servers, which reduces the computation load.
It is also worth noting that servers and workers are actually deployed in the same GPU machines physically, so the alternative worker and server computation maximizes the GPU usage.



\vspace{-0.2cm}
\subsection{DICM implemented by AMS}


As shown in Fig.~\ref{Networks}, DICM can be efficiently trained with AMS. The embeddings of sparse ID features and embedding model are running in servers as design. MLP and Aggregator (detailed in following section) are running in workers.


The distributed GPU training architecture equipped with AMS makes daily updating model with tens of days log data come true, which is critical for real advertising system.
Table~\ref{tab:time} depicts the training time of our best configured model with 18 days of data with different number of GPU.
It is noted that our system has desirable nearly linear scalability with GPUs. We use 20-GPU for a reasonable trade-off between efficiency and economy.

\vspace{-0.1cm}
\begin{table}[!h]
\centering
\begin{tabular}{|c||c|c|c|c|}
  \hline
  \#GPU                           &    5  &  10 & 20 & 40\\ \hline
  Time(h)                       &   62.9    & 32.0    & 17.4   &  10.2  \\ \hline
\end{tabular}
\caption{Training time with different number of GPU}
\label{tab:time}
\vspace{-0.6cm}
\end{table}

\subsection{Inference and online deployment}

The efficiency is crucial for online deployment of the CTR model in large industrial advertising system.
For CTR models with sparse ID features, \emph{e.g.} Embedding\&MLP, ID embeddings are globally placed in key-value storage.
And parameters of MLP part are stored locally in a ranking server.
For each request, the ranking server pulls the ID embeddings and feed them through MLP to obtain the predicted CTR.
This scheme is proved to be of high throughput and low latency in the production environment.

When involving images, especially large number of behavior images, extracting image features could bring heavy computation and communication load.
The embedding model is separable from basic net. Benefitting from that the image embeddings can be computed in advance.
So ranking servers are able to predict DICM efficiently with little modification.
Note that the newly involved images can be embedded by the embedding model, which alleviates the cold start problem of ID features.
DICM increases the response time over the baseline just in a tolerable degree, from 21 milliseconds to 24 milliseconds for each PV request.

\section{Image based user modeling} \label{ch_usernet}



\subsection{Image embedding model}

The embedding model is designed to extract the pixel-level visual information to semantic embedded vector.
Recent progress in computer vision shows that the learnt semantic features for classification tasks have good generalization ability ~\cite{he2015delving, Simonyan2014Very}.
Our empirical studies show that VGG16~\cite{Simonyan2014Very} performs better than trivial end-to-end training from scratch in our application.
But due to the unsatisfactory complexity of VGG16, we adopt a hybrid training: the whole net is split into a fixed part followed by a trainable part that is end-to-end trained with CTR model.


For the fixed part, we adopt the first 14 layers of pre-trained VGG16 net~\cite{Simonyan2014Very}, specifically, from Conv1 to FC6, which generates a 4096-D vector.
It is a careful trade-off between efficacy and efficiency for practical application. E.g., replacing FC6 with 4096-D output by VGG16 FC8 with 1000-D output as the fixed part will lead to 3\% relative performance loss in our experiments.
It indicates that the information reduction in fixed part needs to be controlled, and the input size and the trainable part jointly learnt with the whole network are crucial.
However, when we use lower layer in VGG16 as the fixed part, the computation load in training becomes high and we found the improvement is not significant.
Finally VGG16 FC6 with 4096-D output is selected as the fixed part. 
For the trainable part, a 3-layer fully connected net (4096-256-64-12) is used and outputs a 12-D vector. 

\vspace{-0.2cm}
\subsection{User behavior image aggregator} \label{ch:candidate_aggregator}

For CTR prediction with Embedding\&MLP model, a compact representation of the user is critical.
We need to aggregate the various user data, especially historical behaviors of variable number, to a fixed length vector.
Thus an aggregator block is designed to aggregate numerous behavior image embeddings for this sake.




In fact, similar tasks are involved in many classical problems.
For traditional image retrieval/classification, local features, \emph{e.g.} SIFT ~\cite{lowe1999object}, in an image are aggregated. Classical methods including VLAD~\cite{jegou2010aggregating} and sparse coding ~\cite{yang2009linear} accomplish this with sum or max operation.
For neural machine translation, the context vector for different length sentence is abstracted with recent attentive method~\cite{bahdanau2014neural,Vaswani2017attention}. 
We follow these ideas and explore various designs, especially the attentive method.
Further ID feature information is concerned to propose the Multiple Query Attentive Pooling.





The most straightforward method is to \textbf{concatenate} all behavior image embeddings together, and pad or truncate to a specified length.
But it would suffer a loss when the behavior number is large or when the behavior order changes.
\textbf{Max} and \textbf{sum pooling} are another two direct methods, which can not focus appropriately for diverse user behaviors.
Recently DIN~\cite{zhou2017deep} introduces attentive mechanism to user modeling.
It adaptively captures the most relevant behaviors depending on the ad under consideration.
We also employ this method, and considering the visual relevance, we use ad image as query in attention.
We call this method \textbf{AttentivePooling}.
These methods are illustrated in Fig.~\ref{Aggregator}.

Interactions between different types of features are important.
For example, the category id "T shirt" of ad and the "T shirt" image in user behaviors can be connected, and hence it captures the user's preference to such items better.
Therefore, we propose the \textbf{MultiQueryAttentivePooling} (Fig.~\ref{Aggregator}d) which incorporates both images and IDs for attentive weights generation.
In detail, we design two attentive channels which involve ad image feature and ID feature as queries respectively.
Both of the attentive channels generate their own weights as well as the weighted sum vectors separately, which are then concatenated.
Note that different from the multi-head technique~\cite{Vaswani2017attention}, the MultiQueryAttentivePooling uses different queries for each attentive channel and thus explores distinct relevances with complementarity. 

We empirically compare these aggregator designs in Sec.~\ref{AggregatorSubsection}. 

\section{DICM for \emph{Pre-rank}} \label{ch_prerank}

\begin{figure}
  \centering. 
  \includegraphics[width=0.8\linewidth]{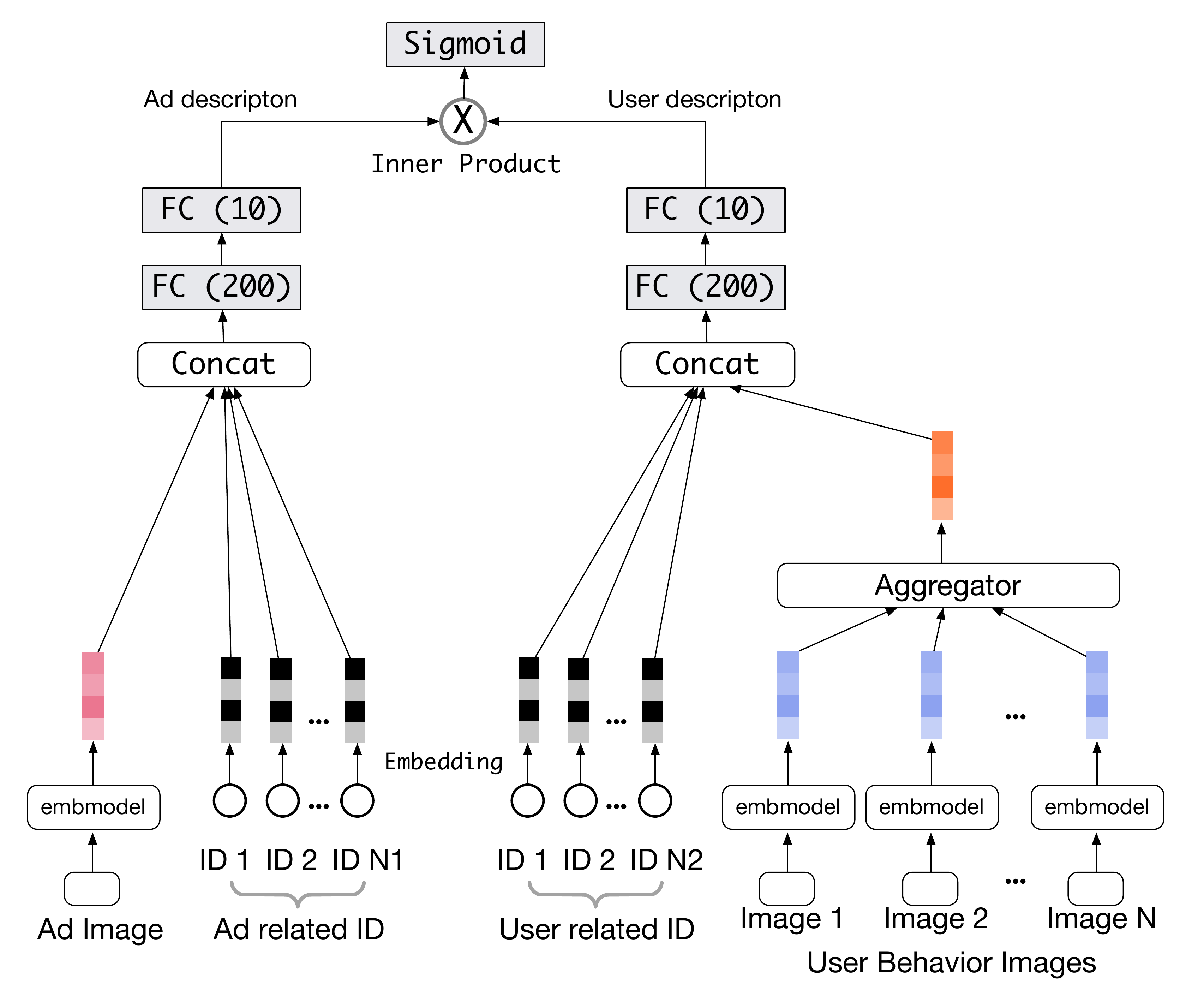}
  \vspace{-0.4cm}
  \caption{DICM network architecture for \emph{Pre-rank}}\label{NetworksPreranking}
  \vspace{-0.3cm}
\end{figure}

DICM framework can be smoothly applied to \emph{Pre-rank} phase that is introduced in \ref{ch_display_adv_sys}.
To speed up online serving, we design the architecture like DSSM ~\cite{huang2013learning} structure which is widely used in efficiency sensitive cross domain search/recommendation tasks.
As shown in Fig.~\ref{NetworksPreranking}, ad and user representations of equal length are first modeled separately with their own features.
As in \emph{Rank}, ID features and images are adopted and embedded with embedding models.
To avoid early fusion of ad and user features, sum pooling is used as aggregator for behavior images.
The final CTR is predicted by an inner product of them.



\section{Experiments}

\subsection{Dataset and evaluation metric} \label{experiment_setting}

The experiment data comes from Taobao's display advertising system. In detail, we construct a closed dataset with log data from arbitrary 19 consecutive days in July 2017.
We use the data of the first 18 days as training set and that of the subsequent day as test set.
In total, the dataset has 3.9 billion training samples and 219 million test samples.
All offline experiments are conducted with this dataset. We use 27 categories of ID features including user profiles, user behaviors, ad description and scenario description, which are simplified from the highly optimized online configuration. The unique user number is about 0.13 billion.
    About 1\% ads are newly uploaded or updated every day.





 
For offline evaluation metric, we adopt AUC (Area Under ROC Curve) which is commonly used in advertising/recommender system~\cite{WideandDeep, mo2015image, chen2016deep}. Besides, we also use the Group AUC (GAUC) introduced in ~\cite{Zhu2017Optimized, zhou2017deep}. 

GAUC is a weighted average of AUC over all users. The GAUC is formulated as:
\begin{equation*}\label{GAUC}
    GAUC = \frac{\sum_{i}\#impression_i * AUC_i}{\sum_{i}\#impression_i}
\end{equation*}
where $\#impression_i$ and $AUC_i$ are the number of impression and AUC corresponding to the i-th user, respectively. In real advertising system, GAUC is proved to be more effective to measure the performance than AUC or Cross Entropy Loss\cite{Zhu2017Optimized}, since the system is personalized and focuses on prediction for each user.

\subsection{Training details}\label{ch_detail}



To speed up training and reduce storage cost, we follow the common-feature technique~\cite{zhou2017deep,chen2016deep,gai2017learning}. In detail, we put together samples corresponding to identical users to form sample groups which share user relevant features as common-features, following~\cite{gai2017learning}.

To describe user behavior, we select the \emph{click} behavior of one specific user in the past 14 days. Since raw data from real system is noisy, we select typical click behavior with reasonable long visiting elapse time. We empirically find such filtering strategy achieves better performance. The average user's behaviors are filtered from more than 200 to 32.6.


We use PReLU~\cite{he2015delving} as the activation for each layer since we empirically find its superiority. Adam~\cite{kingma2014adam} is adopted as parameter optimizer, with the learning rate initialized to 0.001 and decayed by 0.9 after each 24,000 sample batches. The model converges after 2 epochs (128K iterations in our scenario).

\paragraph{\textbf{Partial warm-up}}
Parameter initialization is widely used. Benefitting from the daily update scheme of our system, we can use the trained model of last day as initialization without any extra cost. It is observed that each part of DICM converges at different speed.
The ID embedding is prone to get overfitting due to the sparsity of ID and the large parameter size, while image embedding model requires sufficient training to capture the highly non-linear relation between visual information and user's intention.
So we propose the \emph{partial warm-up} technique. In detail, we use pre-trained (but with training data of different date) model as initialization of all parts except ID embedding (\emph{i.e.} image embedding model, extractor and MLP part), and randomly initialize the ID embedding part.






\subsection{Efficiency study of AMS}

We first study the efficiency superiority of AMS in our application. In detail, we compare with the following two possible ways to save the involved images as follows:

\begin{itemize}
\item \textbf{ store-in-worker.} Storing the images in the worker nodes, along with other training data. 
\item \textbf{store-in-server.} Storing the images in the server nodes as the global dataset and fetching them in key-value format(without sub-model in server).
\end{itemize}

To give the quantitative results, we summarize our typical scenario. There are in total 3.9 Billion training samples processed by a 20-node GPU cluster. For each training iteration, the mini-batch is set to 3000 per node thus the effective minibatch size is 60000. In each sample, the user is related to 32.6 behavior images on average. Benefitting from common-feature technique (\emph{c.f.}~\ref{ch_detail}), each effective minibatch involves about 320k images as well as 1.4 million IDs (excluding images' IDs) according to statistic. There are in total 0.12 billion unique images involved in training, and each one is pre-processed to 4096-D float feature as training input. 

We compare the AMS with the two replacements in Table~\ref{tab:ams_efficiency}. We see the AMS achieves nice system efficiency, while the \emph{store-in-worker} and \emph{store-in-server} strategies suffers major weakness \emph{w.r.t.} storage or communication load. In detail, \emph{store-in-worker} requires 31 times larger storage than that of AMS(5.1G \emph{v.s.} 164M); while \emph{store-in-server} costs 32 times more communication than that of AMS(5.1G \emph{v.s.} 158M).    

\begin{table}[htbp]
\centering
\begin{tabular}{c cc cc}

\toprule
   \multirow{2}{*}{Strategy}  & \multicolumn{2}{c}{Storage}  & \multicolumn{2}{c}{Communication} \\ 
             &   Worker         & Server          &   All     & Image \\ \midrule
   store-in-worker &  5.1G(332T)     & 0      &   128M    & 0   \\ 
   store-in-server &  134M(8.8T)      & 30.3M(2T)       &   5.1G  &  5.0G \\ 
   AMS       &  134M(8.8T)      & 30.3M(2T)       &   158M    & 30M  \\ \bottomrule

\end{tabular}
\caption{Efficiency study of AMS. "Storage" denotes the storage requirements of worker or server group for saving both image and ID data. "Communication" denotes the communication load between worker and server nodes to communicate all data (denoted by"All") and only image data (denoted by"Image"). The listed numbers are the average data size (in Bytes) for each mini-batch of the whole cluster, and these in brackets are the summation of the whole training. 
}
\vspace{-0.3cm}
\label{tab:ams_efficiency}
\end{table}


\vspace{-0.6cm}
\subsection{Ablation studies}\label{AggregatorSubsection}

We first individually study various design details of our approach with offline experiments in this section. For fair comparison, partial warm-up strategy is disabled for all ablation studies unless otherwise specified. 

\paragraph{\textbf{Baseline}}
We set our baseline model for all offline experiments as the Embedding\&MLP model with only sparse ID features, as shown in Fig.~\ref{Networks}, which is a simplified version of the production model in Taobao's display advertising system for clarity.
Note that two special ID fields are also employed as sparse features in baseline: the IDs of ad image and the IDs of user behavior images.
These two ID fields are essential for a fair comparison,
because image features in fact can play a partial role of IDs and we should keep a common basis for both models to show a clean improvement of image semantic information.
Besides, we adopt adaptive regularization~\cite{zhou2017deep} to tackle the overfitting problem of ID features.

\paragraph{\textbf{Study on image information}}
DICM integrates both user behavior images and ad images.
In this section, we conduct an ablation study of the effectiveness of them.
To this end, we start from the baseline and separately use ad images, behavior images and both of them.
Table~\ref{tab:image_comparison} depicts the results on the offline dataset.
It is observed that either behavior images or ad image will boost the baseline, showing the positive effect by introducing visual feature in user and ad modeling.
Further, jointly modeling with both behavior images and ad image will significantly improve the performance.
It is worth noting that the joint gain is much larger than the sum of gains brought by them individually, \emph{i.e.}, 0.0055 \emph{v.s.} 0.0044 in GAUC and 0.0037 \emph{v.s.} 0.0024 in AUC. 
This result strongly indicates the cooperative effect of modeling users and ads by visual information, a desirable effect brought by our DICM.


\begin{table}[htbp]
\centering
\begin{tabular}{l c c c c}
  \toprule
  
   Method          & GAUC   & GAUC gain & AUC     & AUC gain  \\  \midrule
   baseline        &0.6205 & -        &0.6758  & -         \\ 
   ad image        &0.6235 & 0.0030   &0.6772  & 0.0014    \\
   behavior images &0.6219 & 0.0014   &0.6768  & 0.0010    \\
   \textbf{joint}  &\textbf{0.6260} & \textbf{0.0055}    &\textbf{0.6795}  & \textbf{0.0037}    \\
      
\bottomrule   
\end{tabular}
\caption{Comparison of behavior images and ad image, and their combination in DICM.} 
\vspace{-0.5cm}
\label{tab:image_comparison}
\end{table}

\paragraph{\textbf{Study on behavior image aggregator}}
We detail the effects of different aggregators described in Sec.~\ref{ch:candidate_aggregator} with which the behavior image embeddings are exploited in the model.
The results are shown in Table~\ref{tab:aggregator}.
The observations are 3 folds:
i) Concatenation is not appropriate for behavior aggregation, providing inferior performance; sum/max pooling give reasonable improvements.
ii) AttentivePooling shows a remarkably gain with ad images as attention queries.
iii) MultiQueryAttentivePooling brings best results, benefitting from interactions between sparse ID and semantic information in images.


\begin{table}
	\centering
	\begin{tabular}{l c}
		\toprule
		Aggregator     &   GAUC   \\ \midrule
		baseline	& 0.6205	\\
		Only ad images  & 0.6235 \\
		Concatenation        & 0.6232 \\  
		MaxPooling     & 0.6236 \\ 
		SumPooling     & 0.6248 \\
		AttentivePooling & 0.6257 \\
		\textbf{MultiQueryAttentivePooling}&\textbf{0.6260}\\
		\bottomrule
		
	\end{tabular}
	\caption{Result of different aggregator. Aggregators are investigated jointly with ad image.}
	\vspace{-0.5cm}
	\label{tab:aggregator}
\end{table}

\begin{figure}
  \centering
  \includegraphics[width=0.95\linewidth]{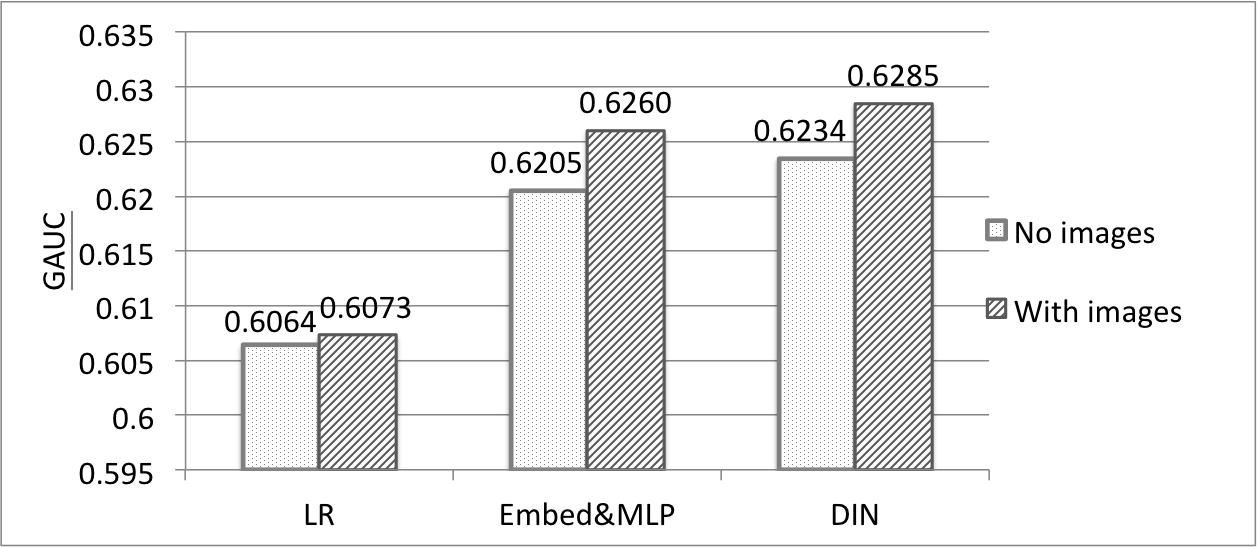}\\
  \caption{GAUC of model with different basic structures}
  \vspace{-0.2cm}
  \label{fig:structure}
  \vspace{-0.3cm}
\end{figure}

\paragraph{\textbf{Study on different basic structure}}
Our work focuses on enhancing CTR prediction model with jointly involving visual information of user behaviors and ads.
The basic network structure design for traditional sparse features is not the central topic in this paper.
We assume that DICM can apply to different basic networks and bring consistent improvement with image features.
To verify it, we test DICM with the classical Logistic Regression (LR) model and recently proposed DIN~\cite{zhou2017deep} model, along with the baseline Embedding\&MLP as basic model.
Fig.~\ref{fig:structure} compares offline metric GAUCs of these models.
It can be seen that models with images consistently outperform their counterpart with only ID features as expected.
DIN with image features performs the best, and largely surpasses classical DIN.
The improvement of LR when enhancing with images is not as much as others.
It is because LR can not fully utilize the high level semantic information of images.


\paragraph{\textbf{Study on partial warm-up}}
We empirically study warm-up strategies by comparing the strategy of non warm-up, partial warm-up and full warm-up.
As shown in Table~\ref{tab:warm-up}, partial warm-up performs best.
Full warm-up leads to the worse result due to a severe overfitting in the ID embedding parameters.

\begin{table}[htbp]
\centering
\begin{tabular}{c c c c}
  \toprule
  warm-up strategy &    Non    &  \textbf{Partial} & Full\\ \midrule
  GAUC     &  0.6260  & \textbf{0.6283}   & 0.6230  \\ 
  \bottomrule
\end{tabular}
\caption{Comparison of warm-up strategy}
\vspace{-0.5cm}
\label{tab:warm-up}
\end{table}

\subsection{Results of DICM}

In this section, we compare the best configured DICM using partial warm-up strategy and MultiQueryAttentivePooling with the baseline by offline metric.
Online A/B test is also conducted and shows a significant improvement over the state-of-the-art in production.

\begin{figure}[htbp]
	\centering
	\includegraphics[width=0.9\linewidth]{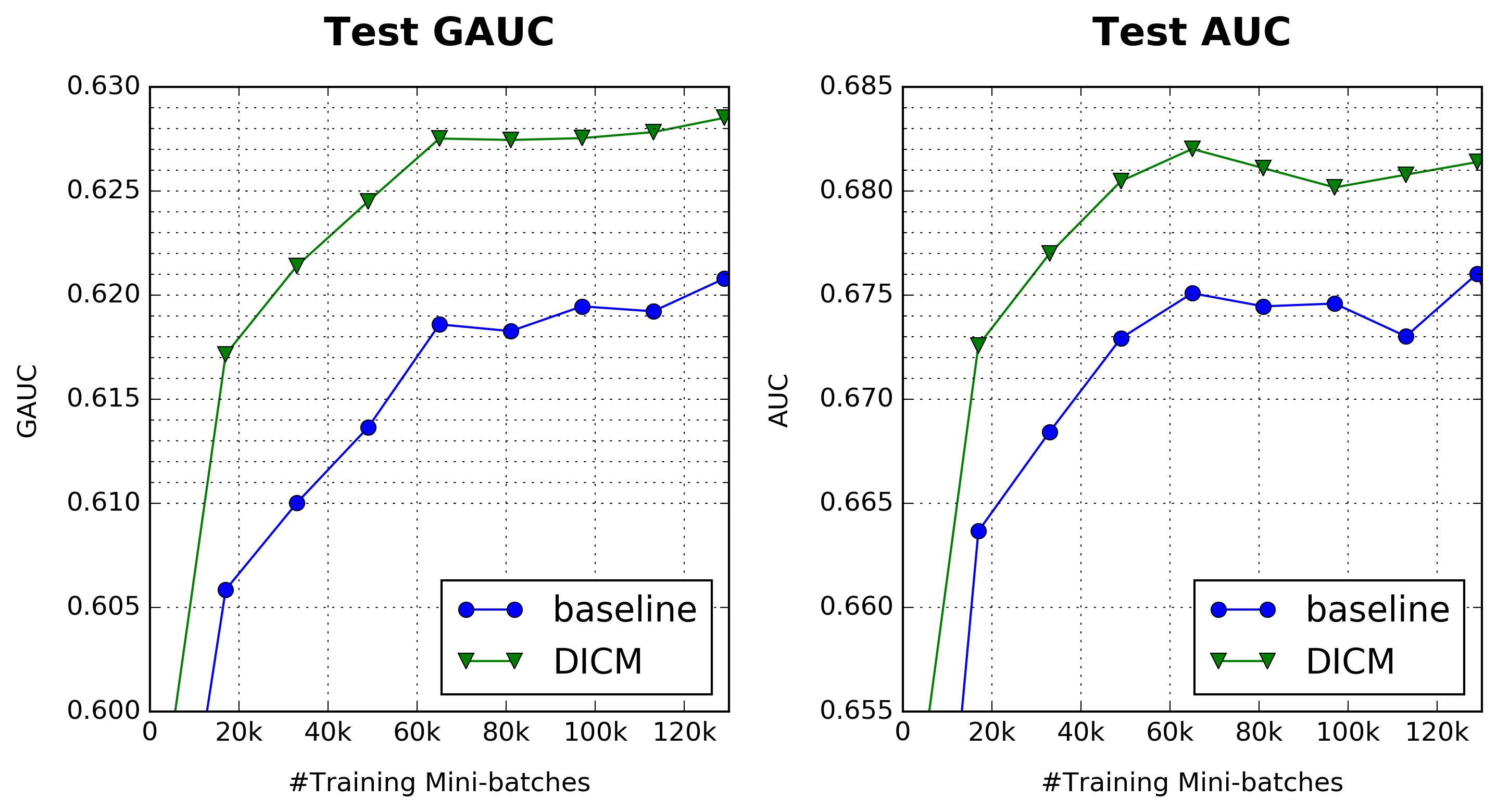}
	\vspace{-0.2cm}
	\caption{Offline result}\label{main_result}
	\vspace{-0.4cm}
\end{figure}

\paragraph{\textbf{Offline results}}
We first evaluate our DICM model with offline dataset.
Partial warm-up strategy and MultiQueryAttentivePooling are adopted.
Table~\ref{tab:main_result} and Fig.~\ref{main_result} show the AUC/GAUC comparison between baseline and the best configured DICM.
DICM outperforms the baseline by 0.0078 GAUC and 0.0055 AUC, which actually are significant improvements in the real system.
Moreover, it is noted from Fig.~\ref{main_result} that the gap between baseline and DICM is consistent during the training process, which indicates the robustness of our approach.

\begin{table}[htbp]
	\centering
	\begin{tabular}{l l l}
		\toprule
		
		Method               & GAUC   & AUC     \\  \midrule
		baseline             & 0.6205($\pm$0.0002) & 0.6758($\pm$0.0003) \\ 
		\textbf{DICM}        & \textbf{0.6283($\pm$0.0002)} & \textbf{0.6814($\pm$0.0003)}    \\ \hline
		Absolute gain       & 0.0078 & 0.0055 \\ 
		\bottomrule
		
	\end{tabular}
	\caption{Offline result, averaged by 5 runs}
	\vspace{-0.6cm}
	\label{tab:main_result}
\end{table}

\paragraph{\textbf{Online A/B test}}

In online A/B test, we set the baseline as the production model, which is the state-of-the-art of our production environment. The production model is extended from the basic net by adding sophisticated cross-product features, such as user age $\times$ ad category, user gender $\times$ weekday \emph{etc.} To be fair, we extend DICM with the same features to get the online version of DICM. The comparison is conducted between the online version of DICM and the production model.
Three key metrics of advertising system are considered: the CTR, eCPM and gross merchandise value per mile (GPM).
As listed in Table~\ref{tab:online}, the DICM achieves a consistent gain in the online A/B test during a 7-day long statistic period.
And these improvements are also remarkable because they show that DICM brings 9.2\% more product impressions and 5.9\% more sales for advertisers and 5.7\% more revenue for the platform.
Considering Taobao's massive volume of traffic and merchandise, the commercial value of DICM is significant.
DICM has been now deployed in Taobao's display advertising system, serving the main traffic for half a billion users and millions of advertisers.

\begin{table}[htbp]
	\centering
	\begin{tabular}{c c c c}
		\toprule
		Date   &  CTR      &  eCPM & GPM\\ \midrule
		Day1           &  +10.0\%  &  +5.5\%  & +3.3\% \\
		Day2           &  +10.0\%  &  +6.8\%  & +8.0\% \\
		Day3           &  +9.1\%   &  +6.6\%  & +1.8\% \\
		Day4           &  +9.9\%   &  +4.8\%  & +7.9\% \\
		Day5           &  +8.2\%   &  +5.0\%  & +2.7\% \\
		Day6           &  +8.2\%   &  +5.4\%  & +9.9\% \\
		Day7           &  +9.0\%   &  +5.7\%  & +8.0\% \\ \hline
		Average        &  9.2($\pm$0.7)\%    &  5.7($\pm$0.7)\%     &5.9($\pm$4.0)\%\\ 
		\bottomrule
	\end{tabular}
	\caption{Relative increments of DICM in online A/B test, counting in 7 consecutive days (Nov 21st-27th, 2017)}
	\vspace{-0.8cm}
	\label{tab:online}
\end{table}

\subsection{Applying to \emph{Pre-rank}}


Finally, we evaluate the performance of applying DICM in \emph{Pre-rank} phase.
The network described in Fig.~\ref{NetworksPreranking} is trained on the offline dataset. As shown in Table~\ref{tab:main_result_preranking}, our DICM again significantly outperforms baseline under both GAUC and AUC metrics. Such  results indicate the prospect of generalizing our framework to other CTR prediction task of advertising/recommendation system.



\begin{table}[htbp]
\centering
\begin{tabular}{l c c c c}
  \toprule
   Method              & GAUC    & GAUC gain  & AUC     & AUC gain\\  \midrule
   baseline            & 0.6165  & -        & 0.6730 & -\\ 
   \textbf{DICM}       & \textbf{0.6225} & \textbf{0.0060}    & \textbf{0.6771} & \textbf{0.0041}\\ 
 \bottomrule
   
\end{tabular}
\caption{Result of DICM for \emph{Pre-rank}}
\label{tab:main_result_preranking}
\end{table}
\vspace{-0.8cm}

\section{Conclusion}


In this paper, we have proposed a novel and efficient distributed machine learning paradigm called AMS. 
Benefitting from it, we manage to utilize massive number of behavior images in capturing user interest for CTR prediction in display advertising. 
We design a complete architecture named DICM that jointly learns ID and visual information for both user and ad description, and depict its superiority via  offline and online experiments.  
Since user behaviors usually incorporate abundant cross-media information such as comment texts, detailed descriptions, images and videos, 
we believe that our proposed AMS and the model study can also benefit future work in this direction.




\bibliographystyle{ACM-Reference-Format}
\bibliography{imgSch}

\end{document}